\documentclass{article}

\usepackage{arxiv}

\usepackage[utf8]{inputenc} 
\usepackage[T1]{fontenc}    
\usepackage{hyperref}       
\usepackage{url}            
\usepackage{booktabs}       
\usepackage{amsfonts}       
\usepackage{nicefrac}       
\usepackage{microtype}      
\usepackage{lipsum}
\usepackage{microtype}
\usepackage{graphicx}
\usepackage{subfigure}
\usepackage{relsize}
\usepackage{amsmath}
\usepackage{amsfonts}
\usepackage{amsthm}
\usepackage{amssymb}
\usepackage{bbm}
\usepackage{natbib}
\usepackage{mathtools}
\usepackage{booktabs} 
\usepackage{changepage}

\usepackage{enumitem}
\usepackage{caption} 

\title{High-Fidelity Vector Space Models \\ of Structured Data}

\author{ 
Maxwell Crouse \thanks{send correspondence to mvcrouse@u.northwestern.edu} \\
  Qualitative Reasoning Group \\
  Northwestern University \\
   Evanston, IL \\
  \texttt{mvcrouse@u.northwestern.edu} \\
   \And
   Achille Fokoue \\ 
   IBM Research AI \\
   Yorktown Heights, NY \\
   \texttt{achille@us.ibm.com} \\
   \And
   Maria Chang \\
   IBM Research AI \\
   Yorktown Heights, NY \\
   \texttt{Maria.Chang@ibm.com} \\
   \And
   Pavan Kapanipathi \\
   IBM Research AI \\
   Yorktown Heights, NY \\
   \texttt{kapanipa@us.ibm.com} \\
   \And
   Ryan Musa \\
   IBM Research AI \\
   Yorktown Heights, NY \\
   \texttt{ramusa@us.ibm.com} \\
   \And
   Constantine Nakos \\
   Qualitative Reasoning Group \\
   Northwestern University \\
   Evanston, IL \\
  \texttt{cnakos@u.northwestern.edu} \\
  \And
   Lingfei Wu \\
   IBM Research AI \\
   Yorktown Heights, NY \\
   \texttt{wuli@us.ibm.com} \\
   \And
   Kenneth Forbus \\
   Qualitative Reasoning Group \\
   Northwestern University \\
   Evanston, IL \\
   \texttt{forbus@northwestern.edu} \\
   \And
    Michael Witbrock \\
   IBM Research AI \\
   Yorktown Heights, NY \\
   \texttt{witbrock@us.ibm.com} \\
}

\begin{document}
\maketitle

\begin{abstract}
Machine learning systems regularly deal with structured data in real-world applications. Unfortunately, such data has been difficult to faithfully represent in a way that most machine learning techniques would expect, i.e. as a real-valued vector of a fixed, pre-specified size. In this work, we introduce a novel approach that compiles structured data into a satisfiability problem which has in its set of solutions \textit{at least} (and often \textit{only}) the input data. The satisfiability problem is constructed from constraints which are generated automatically \textit{a priori} from a given signature, thus trivially allowing for a bag-of-words-\textit{esque} vector representation of the input to be constructed. The method is demonstrated in two areas, automated reasoning and natural language processing, where it is shown to produce vector representations of natural-language sentences and first-order logic clauses that can be precisely translated back to their original, structured input forms.
\end{abstract}

\section{Introduction}

How best to represent structured data such that it is compatible with machine learning algorithms is an open problem. When support vector machines were the predominant approach of the machine learning community, the representations of choice for ordered trees were that of tree-kernels, e.g. \cite{moschitti2006making, rieck2010approximate, moschitti2008tree}. With the rise of deep learning, tree-structured LSTMs have become a popular method to capture structural characteristics of natural language \cite{tai2015improved,miwa2016end}. These approaches are all aimed at combining the expressive power of structured information that is inherent in naturally occurring data (such as predicate-argument structures in natural language and hierarchical spatial information in images) with robust machine learning methods. However, the transformations applied to structured data in order to adapt them to various machine learning models often make the resulting real-valued vector representations opaque and difficult (or impossible) to interpret. Furthermore, automatically learning these representations has the additional challenge of cultivating and annotating the appropriate training data.

In this work, we describe a novel, implemented method for translating arbitrary structured data into vectors of a fixed, pre-determined length. In particular, this method can be applied to any data representable as (potentially disconnected) directed-acyclic graphs containing both ordered and unordered nodes (e.g. sequences, trees, forests, etc.). This method is not a model that is learned from data, but rather is a means of decomposing an input into a set of constraints that were generated \textit{a priori} from a given signature. The method allows for round-trips in the sense that it is possible to encode data into vectors and to decode vectors into their source data. As such, our method has three main advantages when compared to existing approaches. First, our characterization of graphs as a set of constraints is based on first principles, meaning that it does not require training data. Second, there is no opacity in how representations are constructed, thus, we need not guess what a vector represents as we might with a neural model. Third, the representations of our method leave little on the table; that is, the vectors produced by our method can be made to encode everything needed to reconstruct \textit{only} their original, structured input forms.

\section{Related Work}

\subsection{Automated Reasoning}

The work of \cite{kaliszyk2015efficient} provides a summary of previously proposed methods for encoding first-order formulae as feature vectors. In addition, they introduce the use of a discrimination tree (containing all terms seen in a corpus of proofs) as a means of constructing feature representations of an input clause. In \cite{jakubuuv2017enigma}, the features of a clause were defined to be triples (though their method could be extended to larger tuples) formed from outwards path walks of length 3 through the tree-structure of a clause. \cite{kuhlwein2013mash} extracts features from terms in a way that preserves ordering information. Their method takes as input the literals of a clause and extracts patterns up to a maximum depth that encode at least all non-variable symbols and their relative ordering. Each of the aforementioned works share the same flaws, which are the inability to reconstruct the input set of literals from a set of features and the inability to represent elements of a clause past a fixed depth.

\subsection{Natural Language Processing}
Representing language as vectors is a fundamental component of modern natural language processing, making it a natural application for our approach. \cite{brown1992classbased} introduced the popular Brown clustering technique, a greedy algorithm for mapping words to binary vectors; the vectors for semantically- or syntactically-similar words are encouraged to share a prefix, allowing larger clusters to be formed by truncating the vectors. At the level of a sentence, \cite{toutanova2004leaf} gives a method for computing features for syntactic parse trees. They define a new type of tree-kernel, the leaf-projection path, which could be used to extract features from a syntactic parse tree. Leaf-projection paths are paths from leaf nodes to the root of a tree, which they apply to the task of parse selection. Like most others, their method discards ordering information; thus the process of translating from a set of features back to their original tree would be difficult. More recent work by \cite{socher2013recursive} and \cite{tai2015improved} demonstrates how embedding sentences within a continuous vector space by recursing over their parse trees can improve performance on sentence classification tasks like sentiment recognition and semantic similarity. These models are designed to capture the compositional nature of language more explicitly than recurrent neural networks such as LSTMs \cite{hochreiter1997lstm}. 

\subsection{Graph Decomposition Methods}

Traditionally, graph kernels perform graph-level classification by comparing sub-structures of graphs. These kernels recursively decompose the graphs into small sub-structures, and then define a feature map over these sub-structures for the resulting graph kernel. Sub-structures including random walks \cite{gartner2003graph,kashima2003marginalized}, shortest paths \cite{borgwardt2005shortest}, cycles \cite{horvath2004cyclic}, subtree patterns \cite{mahe2009graph,shervashidze2009fast}, tree fragments  \cite{zanzotto2012distributed}, and graphlets \cite{shervashidze2009efficient} have been widely used. Conceptually, these notable graph kernels can be viewed as instances of a general kernel-learning framework called R-convolution for discrete objects \cite{haussler1999convolution,shervashidze2011weisfeiler}. However, these graph decomposition methods ignore the order information of the sub-structures, which makes it hard to reconstruct the original inputs from their feature representations. 

\section{Preliminaries}

For our method to deterministically encode a graph as a real-valued vector, the input graph must hold two properties. First, there must be some notion of root nodes. That is, there must be nodes that characterize the beginning of a graph. Second, the input graph must be directed and acyclic.

We define a signature (i.e. all possible nodes a graph can be constructed from) to be the union of two sets, $\Sigma = \Pi \cup \Omega$. The elements of $\Pi$ are root nodes and the elements of $\Omega$ are internal nodes. In application, there need not be a distinction between $\Pi$ and $\Omega$, i.e. $\Pi = \Omega$. However, allowing for such a distinction gives the freedom to construct more compact representations when applicable. We refer to the number of children a node may have as its arity. Importantly, nodes with the same label but different arities are treated as distinct. For most examples, our notation will be that which is commonly seen in mathematical logic, e.g. for a tree with root $f$ and children $a$ and $b$, we write $f(a, b)$. When we refer to sets of root nodes and internal nodes, we will use the uppercase letters $P$ and $F$, where $P^k$ and $F^k$ are used to denote a set of root nodes and internal nodes, respectively, each member of which has arity $k$.

Our approach encodes graphs as satisfiability problems. Given a graph, $G$, which has been generated from a signature $\Sigma$, our approach constructs a satisfiability problem that has a set of solutions which \textit{at least} includes $G$.

\section{Graph Encoding}

Our  method  maintains  two  sets: a set of symbols, \textit{S}, and a set of constraints, \textit{C}. Recall, all graphs are constructed from a signature $\Sigma = \Pi \cup \Omega$. As our goal is to represent any graph with only elements from $\Sigma$, in the simplest case we have $S = \Sigma$. In general, we maintain a distinction between $S$ and $\Sigma$, as we may wish the correspondence between them to be given by a mapping function. The set $\textit{C}$ consists of constraints that each node will be mapped to. We now describe the construction of $C$.

\subsection{Ordered Node Constraints}

Let $\Pi_{ord} \subseteq \Pi$ and $\Omega_{ord} \subseteq \Omega$ be the subsets of root and internal nodes which are ordered. First, all such root and internal nodes are divided into sets by arity, i.e.

\begin{enumerate}
\item[] $P_{ord} = \{ \ P_1, \ldots, P_m \ \mid \ P_i : arity(p) = i, \forall p \in \Pi_{ord} \ \}$
\item[] $F_{ord} = \{ \ F_1, \ldots, F_n \ \mid \ F_i : arity(f) = i, \forall f \in \Omega_{ord} \ \}$
\end{enumerate}

Next, define a function $split$ which takes as input a width $w \in \mathbb{N}$ and a set $M$ and randomly partitions $M$ into at most $w$ equally-sized disjoint sets. Using $split$, we define a base to our constraints as follows:

\begin{enumerate}
\item[] $\overline{P}_{ord} = \mathlarger\bigcup_{P_i \in P_{ord}} split(w, P_i)$

\item[] $\overline{F}_{ord} = \mathlarger\bigcup_{F_i \in F_{ord}} split(w, F_i)$
\end{enumerate}

From $\overline{P}_{ord}$ and $\overline{F}_{ord}$ (both of which are sets of sets) we define the set of all ordered constraints $C_{ord}$ to be

\begin{enumerate}
\item[] $C_P = \mathlarger\bigcup_{P^k \in \overline{P}_{ord}} \{ P^k(F_1, \ldots, F_k) \ \mid \ \forall F_1, \ldots, F_k \in split(w, \Omega) \} $

\item[] $C_F = \mathlarger\bigcup_{F^k \in \overline{F}_{ord}} \{ F^k(F_1, \ldots, F_k) \ \mid \ \forall F_1, \ldots, F_k \in split(w, \Omega) \} $

\item[] $C_{ord} = C_P \cup C_F$
\end{enumerate}

Notice that every constraint, i.e. each $P^k(\ldots)$ and $F^k(\ldots)$, is defined with a separate invocation of $split$.

Each element of $C_{ord}$ should be thought of as a pattern that ground expressions will be matched against. To be more concrete, matchability between an ordered pattern $P^k(F_1, \ldots, F_k)$ and some subgraph $p(f_1(\ldots), \ldots, f_k(\ldots))$ is defined to be

\begin{enumerate}
\item The lead elements match, i.e. $p \in P^k$

\item Each immediate argument matches, i.e. $f_i \in F_i, \forall i$
\end{enumerate}

The same definition holds for subgraphs that begin with internal nodes. Note that the definition of matchability as presented above only considers the root or internal node and the internal nodes of its immediate arguments. Whether or not an immediate argument has arguments itself has no bearing on whether it will match to some pattern.

\subsubsection{Space Requirements}

For an ordered node constraint with arguments drawn from a $w$-way random partitioning of $\Omega$, all combinations of possible arguments are needed. Thus, with a maximum arity of $m$ across all ordered root nodes, a maximum arity of $n$ across all ordered internal nodes, and a width $w$, the number of ordered node constraints is given by
\begin{gather*}
|C_{ord}| = w \mathlarger\sum_{i = 1}^{m} w^{i} + w \mathlarger\sum_{i = 1}^{n} w^{i}
\end{gather*}

\begin{figure}[t]
\centering
    \begin{minipage}{0.45\textwidth}
        \centering
            \includegraphics[scale=0.7]{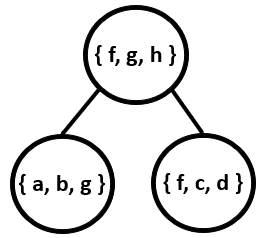}
            \caption{An ordered node constraint that matches $f(a, c)$ but \textit{does not} match $f(c, a)$.}
            \label{ordconstr}
    \end{minipage}\hfill
    \begin{minipage}{0.45\textwidth}
        \centering
            \includegraphics[scale=0.7]{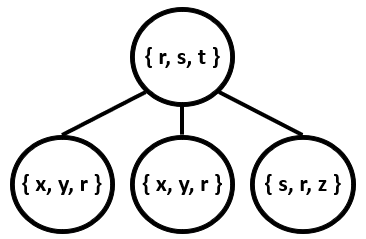}
            \caption{An unordered node constraint that matches $s(x, y, z)$ and $s(z, x, y)$ but \textit{does not} match $s(y, y, x)$}
            \label{unordconstr}
    \end{minipage}
\end{figure}
\subsection{Unordered Node Constraints}

Let $\Pi_{unord} \subseteq \Pi$ and $\Omega_{unord} \subseteq \Omega$ be the subsets of root and internal nodes which are unordered. As before, all root and internal nodes are first divided into sets by arity, i.e.
\begin{enumerate}
    \item[] $P_{unord} = \{ P_1, \ldots, P_m \mid P_i : arity(p) = i, \forall p \in \Pi_{unord} \}$
    \item[] $F_{unord} = \{ F_1, \ldots, F_n \mid F_i : arity(f) = i, \forall f \in \Omega_{unord} \}$
\end{enumerate}

With $split$ defined as before, we construct a base to our set of unordered constraints as follows:

\begin{enumerate}
    \item[] $\overline{P}_{unord} = \mathlarger\bigcup_{P_i \in P_{unord}} split(w, P_i)$
    \item[] $\overline{F}_{unord} = \mathlarger\bigcup_{F_i \in F_{unord}} split(w, F_i)$
\end{enumerate}

Enabling our approach to deterministically map an unordered node and its immediate arguments to a constraint is a bit more involved than in the ordered case. First, our approach associates each element of $\overline{P}_{unord}$ and $\overline{F}_{unord}$ with a unique random ordering of elements from $\Omega$. To do this, define a function $order$, which takes as input a width $w \in \mathbb{N}$, a length $l \in \mathbb{N}$, and a set $M$. $order$ first uses the $split$ function to divide $M$ into $w$ random equally-sized disjoint partitions, producing a set $\overline{M}$

\begin{enumerate}
    \item[] $\overline{M} = split(w, M) = \{ \ M_1, \ldots, M_w \ \}$
\end{enumerate}

From $\overline{M}$, $order$ produces the set of all sequences of length $l$ with elements drawn from $\overline{M}$ such that no $M_i$ is placed before an $M_{j}$ where $i > j$, e.g. for $w = 4$ and $l = 3$ one such sequence is $(M_1, M_1, M_2)$ and another is $(M_3, M_4, M_4)$. With $P_{unord}$, $F_{unord}$, and $order$ defined, we can now construct the set of all unordered constraints $C_{unord}$ 

\begin{enumerate}
    \item[] $C_P = \mathlarger\bigcup_{P^k \in \overline{P}_{unord}} \{ P^k(F_1, \ldots, F_k) \mid \forall (F_1, \ldots, F_k) \in order(w, k, \Omega) \} $
    \item[] $C_F = \mathlarger\bigcup_{F^k \in \overline{F}_{unord}} \{ F^k(F_1, \ldots, F_k) \mid \forall (F_1, \ldots, F_k) \in order(w, k, \Omega) \} $
    \item[] $C_{unord} = C_P \cup C_F$
\end{enumerate}

As before, each element of $C_{unord}$ is to be considered a pattern against which subgraphs will be matched. Consider a particular subgraph, e.g. $p(f_1(\ldots), \ldots, f_k(\ldots))$, and a pattern, $P^k(F_1, \ldots, F_k)$. The subgraph matches the pattern if
\begin{enumerate}
    \item The lead elements match, i.e. $p \in P^k$
    \item The set $\{ f_1, \ldots, f_k \}$ can be put into a one-to-one correspondence with the set $\{ F_1, \ldots, F_k \}$ (where a correspondence between $f_i$ and $F_i$ is given by $f_i \in F_i$)
\end{enumerate}
\subsubsection{Space Requirements}

Consider a length $l$ sequence $s$ with elements drawn from $\{ M_1, \ldots, M_w \}$ that satisfies the constraint that no $M_i$ can come before an $M_j$ where $i > j$. Let $m_i$ be the number of times $M_i$ is found in $s$. For our sequence of length $l$, it must be the case that $m_1 + \ldots + m_w = l$. Each solution to this equation corresponds to a single valid sequence, i.e. the sequence beginning with $m_1$ instances of $M_1$, followed by $m_2$ instances of $M_2$, etc. The total number of solutions to this equation is $\binom{l + w - 1}{l}$, and is given by the Stars and Bars theorem, where $l$ indistinguishable balls (i.e. positions in $s$) are split between $w$ labeled urns (i.e. transitions from $M_i$ to $M_j$ where $i < j$).

With $m$ being the maximum arity across all unordered root nodes, $n$ the maximum arity across all unordered internal nodes, and $w$ being the number of random partitions specified for the $split$ function, the number of unordered constraints is given by
\begin{gather*}
|C_{unord}| = w \mathlarger\sum_{i = 1}^{m} \binom{i + w - 1}{i} + w \mathlarger\sum_{i = 1}^{n}\binom{i + w - 1}{i}
\end{gather*}

\subsection{Parent Constraints}

Parent constraints are generated in much the same way as unordered node constraints. Letting $m \in \mathbb{N}$ be the maximum number of parents a node can have and $w \in \mathbb{N}$ be a width, we first define a set of pairs that will be used to build the parent constraints from
\begin{enumerate}
    \item[] $\overline{F}_{par} = \mathlarger\bigcup_{i = 1}^{m} \  \mathlarger\bigcup_{F \in split(w, \Omega)} \{ \ ( i, F ) \ \}$
\end{enumerate}

Note that, because root nodes cannot have parents, there is no need to include them in this initial step of building the parent constraints. Likewise, we know that leaf-nodes cannot have children, thus we define $\Sigma_{par} \subset \Sigma$ to be the set of all nodes that can be parents (i.e. non-leaf nodes). With $order$ defined as before and $c \in \mathbb{N}$ defined to be the parent width, we construct the set of parent constraints $C_{par}$ as follows 

\begin{enumerate}
    \item[] $C_{par} = \mathlarger\bigcup_{(i, F) \in \overline{F}_{par}} \{ (F, S_1, \ldots, S_i) \mid \forall (S_1, \ldots, S_i) \in order(c, i, \Sigma_{par}) \} $
\end{enumerate}

The conditions for matchability between a node and its parents with a parent constraint are analogous to what was defined for the constraints of $C_{unord}$. For a particular node $f$ with parents $p_1, \ldots, p_i$ and a given pattern, $(F, S_1, \ldots, S_i)$ drawn from $C_{par}$, the conditions for matchability are
\begin{enumerate}
\item The child elements match, i.e. $f \in F$

\item The set $\{ p_1, \ldots, p_i \}$ can be put into a one-to-one correspondence with the set $\{ S_1, \ldots, S_i \}$ (where a correspondence between $p_j$ and $S_j$ is given by $p_j \in S_j$)
\end{enumerate}

\subsubsection{Space Requirements}

The space requirements of parent constraints is given by the same style of proof as with unordered node constraints. Letting $m$ be the maximum number of parents any internal node will have, the number of parent constraints is given by
\begin{gather*}
|C_{par}| = w \mathlarger\sum_{i=1}^{m} \binom{i + w - 1}{i}
\end{gather*}

\begin{figure}[t]
\centering
    \begin{minipage}{0.45\textwidth}
        \centering
            \includegraphics[scale=0.7]{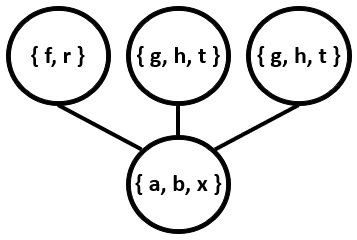}
            \caption{A parent constraint that would match with a node $a$ being the child of three expressions $f(a, c, b)$, $g(d, a)$, and $g(a, c)$.}
            \label{parconstr}
    \end{minipage}\hfill
    \begin{minipage}{0.45\textwidth}
        \centering
            \includegraphics[scale=0.7]{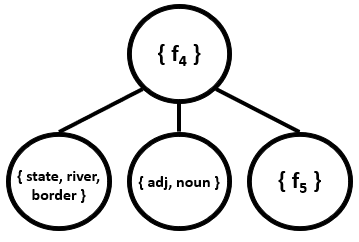}
            \caption{A sequence node constraint that would match to the fourth word of "What is the state with the largest population?"}
            \label{seqconstr}
    \end{minipage}
\end{figure}

\subsection{Sequence Node Constraints}

While sequence constraints are merely a special case of ordered node constraints, we give a distinct description of them here because of how frequently sequences occur in domains of interest, e.g. language processing. To start, the elements of a sequence can be thought of as zero-arity internal nodes, e.g. the word "state". The position of an element in a sequence can then be thought of as a non-zero-arity internal node which takes zero-arity nodes as arguments as well as a link to the next internal node, e.g. the third word of a sentence being "state" is given as $f_3(state, f_4(\ldots))$. We will distinguish between our zero-arity internal nodes (constants) as members of the set $\Omega_{const}$. With a maximum sequence length $l \in \mathbb{N}$, we then have our set of root nodes be $\Pi = \{ f_1 \}$, and the remainder of our internal nodes be $\Omega_{func} = \{ f_2, \ldots, f_l \}$, with $\Omega = \Omega_{const} \cup \Omega_{func}$.

Sequences need not be limited to having one constant per position. We allow every element of the sequence to have a fixed size $t$, e.g. for $t = 4$, one could specify a word, its prefix, its suffix, and its part-of-speech in one position of a sentence. Let $\Omega_{i} \subseteq \Omega_{const}$ be the subset of constants that can be placed at index $i$ of a particular position of a sequence, and let $w \in \mathbb{N}$ be a pre-specified width, then we define $C_{seq}$ as
\begin{enumerate}
    \item[] $C_{P} = \{ \ f_1(split(w, \Omega_1), \ldots, split(w, \Omega_t), \emptyset), \ f_1(split(w, \Omega_1), \ldots, split(w, \Omega_t), f_2) \ \}$
    \item[] $C_F = \bigcup_{j = 2}^{l-1} \{ \ f_j(split(w, \Omega_1), \ldots, split(w, \Omega_t), \emptyset), \ f_j(split(w, \Omega_1), \ldots, split(w, \Omega_t), f_{j+1}) \ \}$
    \item[] $C_L = \{ \ f_l(split(w, \Omega_1), \ldots, split(w, \Omega_t), \emptyset) \ \}$
    \item[]$C_{seq} = C_P \cup C_F \cup C_L$
\end{enumerate}

The conditions for matchability are as one would expect. Consider the $i$-th element $e_i$ (where $e_i$ is a tuple of $t$ elements, i.e. $(e_{i1}, \ldots, e_{it})$) of a sequence $s$ and a constraint $f_j(F_1, \ldots, F_t, l_k)$. Matchability is given by
\begin{enumerate}
    \item  $i = j$
    \item Each argument matches, i.e. $e_{ir} \in F_r, \forall r$
    \item If $i = |s|$ then $l_k = \emptyset$, otherwise $l_k = f_{j + 1}$
\end{enumerate}

\subsubsection{Space Requirements}

With $w$ the number of random partitions, and $s$ the maximum number of entries for each sequence node, the size of $C_{seq}$ is given by
\begin{gather*}
|C_{seq}| = w^s + 2 \mathlarger\sum_{i = 1}^{|\Pi|} w^s + 2 \mathlarger\sum_{i = 1}^{|\Omega_{func}| - 1} w^s
\end{gather*}

\subsection{Vector Space Representations of Structured Data}

Recall, our method maintains two sets, 
\begin{enumerate}
\item[] $S = \Sigma$
\item[] $C = C_{ord} \cup C_{unord} \cup C_{par}$
\end{enumerate}

The decomposition of a given graph $G$ into a multi-set of symbols $\overline{S}$ and constraints $\overline{C}$ is fairly straightforward
\begin{enumerate}
\item[] $\overline{S} = $ every symbol in $G$
\item[] $\overline{C} = $ every constraint that matches a subgraph of $G$
\end{enumerate}

The disjointness enforced by the function $split$ in conjunction with the conditions for matchability described ensure that at each ordered root node of $G$ there will be \textit{exactly} one matchable constraint from $C_{ord}$. Similarly, for ordered internal nodes with arguments, the aforementioned requirements will result in one matchable element from $C_{ord}$. To ensure that unordered nodes and parent relationships are associated with only one constraint from $C_{unord}$ or $C_{par}$, one simply needs to select from the candidate elements of $C_{unord}$ or $C_{par}$ that match with a particular node in a principled fashion. For instance, one might select the constraint for which the one-to-one correspondence between it and the unordered node is lexicographically least with respect to the argument ordering of the unordered node. It is critically important that each node is associated with at most one ordered or unordered node constraint and at most one parent constraint.

A vector representation $V$ can be constructed for $G$ as follows: Assign every symbol and constraint an index in a vector of size $|S| + |C|$. To convert a graph $G$ into a feature vector, simply add the number of times a particular symbol in $\overline{S}$ and constraint in $\overline{C}$ occurs to the appropriate index of $V$. The key property of this vector representation is that it is fixed in length across any input. That is, any graph that is representable with elements from $S$ will be mapped to some vector in $\mathbb{R}^{|S| + |C|}$.

\subsection{Parallel Constraint Sets}

At this point, we have what we need to define a satisfiability problem that has a set of solutions which \textit{at least} includes the original input graph. To enable our approach to create more constrained problems, we add $t$ parallel set of constraints $\{ \ C_1, \ldots, C_t \ \}$, each of which is constructed with the same method described for the initial formulation of $C$. Now, when producing a vector representation of a graph, $\overline{S}$ remains the same, but each $\overline{C_i}$ is the set of patterns applicable to each node at constraint set $i$ in the $t$ parallel sets of constraints. Key to note is the fact that the randomness of $split$ will make each parallel set of constraints completely independent of the others. The vector associated with each graph will now be a member of $\mathbb{R}^{|S| + \sum_i^t |C_i|}$.

\section{Graph Decoding}

Given a vector $v \in \mathbb{R}^{|S| + \sum_i^t |C_i|}$, we wish to reconstruct its source graph, $G$, from $S$ and $C$. For each $j \in [ \ |S| + \sum_i^t |C_i| \ ]$, our approach first extracts $v_j$ number of symbols or constraints (whichever is associated with index $j$). This gives a multi-set of symbols, $\overline{S}$, and a set of multi-sets of constraints, $\overline{C} = \{ \overline{C_1}, \ldots, \overline{C_t} \}$. Each node in $\overline{S}$ is either a root node, an internal node with arguments, or a leaf node.

Recall that each vector represents a satisfiability problem which has in its set of solutions at least the vector's source graph. Thus, we first begin by describing what each propositional variable of the satisfiability problem represents.

Every constraint should be considered a pattern formed by constituent subsets of $S$, e.g. ordered node constraints maintain a set of symbols for the parent as well as sets of symbols for each argument. Consider just one of the parallel sets of constraints for a given input, $\overline{C^\prime} = \overline{C_i}$. With $\overline{C^\prime}_{node} = \overline{C^\prime}_{ord} \cup \overline{C^\prime}_{unord}$, every constituent set of our constraints can be categorized as one of the following classes: 
\begin{enumerate}
    \item \makebox[6.4em][l]{Lead sets} $\{ S \ | \ S(\ldots) \in \overline{C^\prime}_{node} \}$

    \item \makebox[6.4em][l]{Argument sets} $\{ F \ | \ S(\ldots, F, \ldots) \in \overline{C^\prime}_{node} \}$

    \item \makebox[6.4em][l]{Parent sets} $\{ F \ | \ (S, \ldots, F, \ldots) \in \overline{C^\prime}_{par} \}$

    \item \makebox[6.4em][l]{Child sets} $\{ S \ | \ (S, \ldots) \in \overline{C^\prime}_{par} \}$
\end{enumerate}

\begin{figure}[t]
\centering
    \begin{minipage}{0.47\textwidth}
        \centering
            \includegraphics[scale=0.7]{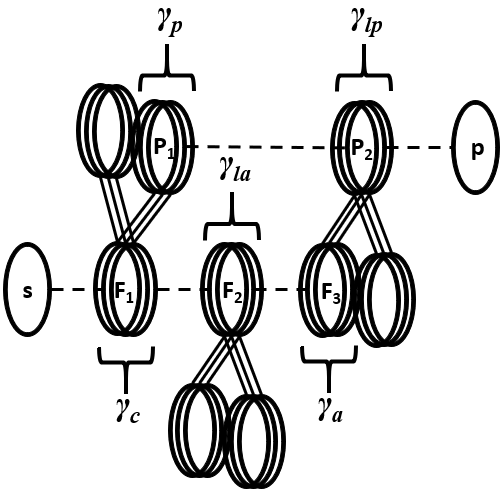}
            \caption{A tuple representing an edge in $G$ when $t = 3$}
            \label{association}
    \end{minipage}\hfill
    \begin{minipage}{0.47\textwidth}
        \centering
            \includegraphics[scale=0.55]{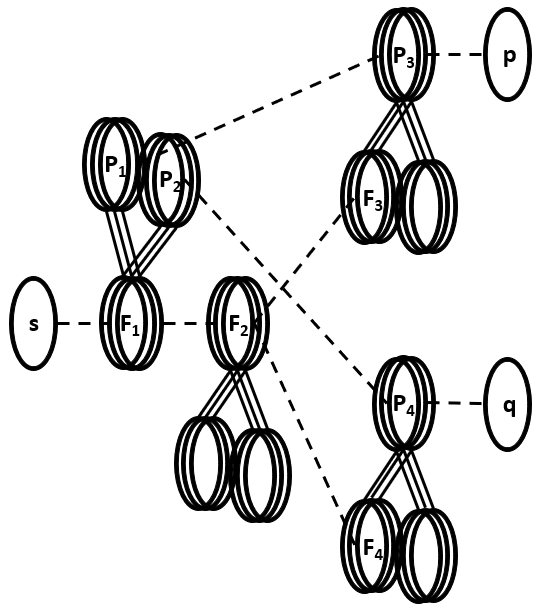}
            \caption{Two (not mutually-exclusive) tuples sharing $s$, $\gamma_c$, $\gamma_{la}$}
            \label{multpar}    
            \end{minipage}
\end{figure}
For every symbol $s \in \overline{S}$, our approach maps over each constituent set type to collect all valid $t$ way associations of constituent sets. More concretely, it collects across all $t$ parallel sets of constraints every combination of constituent sets (with one set from each of the $t$ parallel sets of constraints) such that each set contains $s$, i.e. 
\begin{enumerate}
\item \makebox[5em][l]{Leads} $\Gamma_{l} = \{ (s, S_1, \ldots, S_t) | \forall s \in \overline{S} : s \in S_i \}$

\item \makebox[5em][l]{Arguments} $\Gamma_{a} = \{ (s, F_1, \ldots, F_t) | \forall s \in \overline{S} : s \in F_i \}$

\item \makebox[5em][l]{Parents} $\Gamma_{p} = \{ (s, F_1, \ldots, F_t) | \forall s \in \overline{S} :  s \in F_i \}$

\item \makebox[5em][l]{Children} $\Gamma_{c} = \{ (s, S_1, \ldots, S_t) | \forall s \in \overline{S} : s \in S_i \}$
\end{enumerate}

The elements from each of the $\Gamma$ sets are collected into tuples which will underlie the propositional variables of the satisfiability problem our approach defines. A tuple  $(s, p, \gamma_p, \gamma_c, \gamma_{lp}, \gamma_{a}, \gamma_{la})$ contains seven elements:
\begin{enumerate}
\item \makebox[8.5em][l]{Argument symbol} $s$
\item \makebox[8.5em][l]{Parent symbol} $p$
\item \makebox[8.5em][l]{Parent association} $\gamma_p = (p, F_1, \ldots, F_t) \in \Gamma_{p}$
\item \makebox[8.5em][l]{Child association} $\gamma_c = (s, S_1, \ldots, S_t) \in \Gamma_{c}$
\item \makebox[8.5em][l]{Lead of parent} $\gamma_{lp} = (p, S_1, \ldots, S_t) \in \Gamma_{l}$
\item \makebox[8.5em][l]{Argument of parent} $\gamma_{a} = (s, F_1, \ldots, F_t) \in \Gamma_{a}$
\item \makebox[8.5em][l]{Lead of argument} $\gamma_{la} = (s, S_1, \ldots, S_t) \in \Gamma_{l}$
\end{enumerate}

For a tuple to be valid, two properties must hold. First, the sets in both $\gamma_c$ and $\gamma_{p}$ as well as in $\gamma_{lp}$ and $\gamma_{a}$ must be constituent sets of the same constraints. Second, if the associated constraint of $\gamma_{lp}$ and $\gamma_{a}$ is an ordered constraint, then the elements of $\gamma_{a}$ must all be associated with the same argument position across each of the $t$ underlying ordered constraints. When the argument symbol is a leaf node, $\gamma_{la}$ will be replaced with a skolem placeholder constant. Similarly, when the argument symbol is a root node, $p$, $\gamma_{p}$, and $\gamma_{lp}$ will be replaced with a skolem placeholder constant. Figure \ref{association} provides an example of a tuple.

Each tuple is associated with a propositional variable, i.e. $r = (s, p, \gamma_p, \gamma_c, \gamma_{lp}, \gamma_{a}, \gamma_{la})$, with $\Phi$ being the set of all such variables. Let $\pi$ be a function that takes a constituent set and returns all variables from $\Phi$ that are associated with a tuple containing that set, e.g. for a parent set $P_k$
\begin{enumerate}
\item[] $\pi(P_k) = \{ r \in \Phi \ | \ r = (s, p, \gamma_p, \gamma_c, \gamma_{lp}, \gamma_{a}, \gamma_{la}) : P_k \in \gamma_{p} \}$
\end{enumerate}

We define our first conjunctions as follows
\begin{align*} 
\mathcal{P} & = \mathlarger\bigwedge_{\gamma_{p} \in \Gamma_{p}} \ \ \bigg( \ \ \mathlarger\bigwedge_{F_k \in \gamma_p} \ \ \bigg( \ \ \mathlarger\bigoplus_{r \in \pi(F_k)} \ r \ \ \bigg) \bigg)   \\
\mathcal{A} & = \mathlarger\bigwedge_{\gamma_{a} \in \Gamma_{a}}  \ \ \bigg( \ \  \mathlarger\bigwedge_{F_k \in \gamma_a}  \ \ \bigg( \ \  \mathlarger\bigoplus_{r \in \pi(F_k)} \ r  \ \ \bigg) \bigg) \ \ 
\end{align*}

Now define a function $\phi$ which returns all variables containing the given constituent set or symbol and returns a set of sets of variables, where each internal set of variables share the \textit{exact} same $s$, $\gamma_{la}$, and $\gamma_c$ (Figure \ref{multpar} provides an example of two such tuples).
\begin{align*} 
\mathcal{S} & = \mathlarger\bigwedge_{s \in \overline{S}} \ \  \bigg( \ \  \mathlarger\bigoplus_{R \in \phi(s)} \ \  \bigg( \ \  \mathlarger\bigvee_{r \in R} \ r \ \  \bigg) \bigg) \ \   \\
\mathcal{L} & = \mathlarger\bigwedge_{\gamma_{la} \in \Gamma_{l}} \ \  \bigg( \ \  \mathlarger\bigwedge_{S_k \in \gamma_{la}} \ \  \bigg( \ \  \mathlarger\bigoplus_{R \in \phi(S_k)} \ \  \bigg( \ \  \mathlarger\bigvee_{r \in R} \ r \ \  \bigg) \bigg) \bigg)   \\
\mathcal{C} & = \mathlarger\bigwedge_{\gamma_{c} \in \Gamma_{c}} \ \  \bigg( \ \  \mathlarger\bigwedge_{S_k \in \gamma_c} \ \  \bigg( \ \  \mathlarger\bigoplus_{R \in \phi(S_k)} \ \  \bigg( \ \  \mathlarger\bigvee_{r \in R} \ r \ \  \bigg) \bigg) \bigg)
\end{align*}

\begin{figure}[t]
\begin{center}
\includegraphics[scale=0.6]{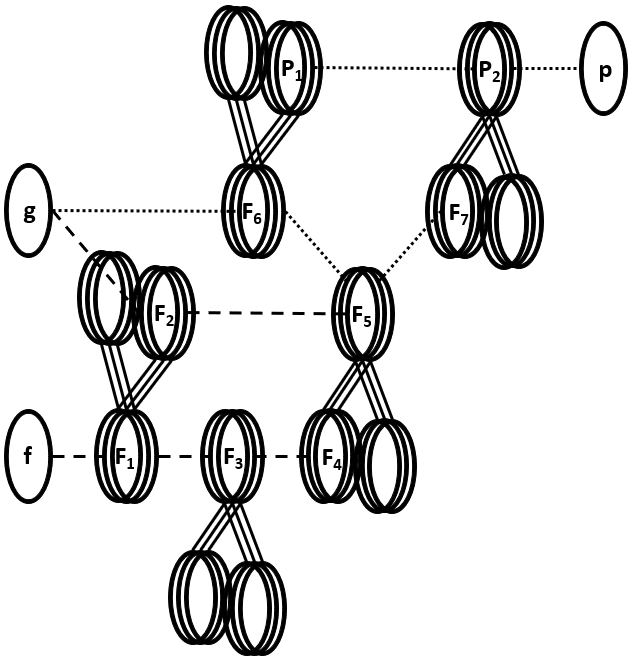}
\end{center}
\caption{A connection between two variables through parallel connectivity}
\label{parconn}
\end{figure}Taking inspiration from the theory of structure-matching presented in \cite{gentner1983structure, forbus2017extending, mclure2015extending}, we make use of multiple sets of constraints through a notion of parallel connectivity. Our specification of parallel connectivity constraints (which differs slightly from that of the aforementioned works) enforces that, if a set of internal nodes across all $t$ partitions match together, then their parents must match together as well (see Figure \ref{parconn}). More concretely, consider a propositional variable $r$ and its associated tuple $(s, p, \gamma_p, \gamma_c, \gamma_{lp}, \gamma_{a}, \gamma_{la})$. Define a function $\sigma$ which takes a propositional variable and returns the set of all variables from $\Phi$ whose lead of argument element, i.e. $\gamma_{la}$, is the \textit{exact} same set as the input propositional variable's lead of parent element, i.e. $\gamma_{lp}$, and whose argument symbol, i.e. $s$, is the \textit{exact} same as the input variable's parent symbol, i.e. $p$. Parallel connectivity constraints are incorporated as implications, where a propositional variable $r$ with $\gamma_{lp}$ and $p$ being true enforces that at least one propositional variable from $\sigma(r)$ is true.
\begin{gather*}
\mathcal{PC} = \mathlarger{\bigwedge_{r \in \Phi}} \ \ \bigg( \ \ r \ \ \mathlarger{\boldsymbol{\Rightarrow}} \ \ \bigg( \ \ \mathlarger\bigvee_{q \in \sigma(r)} \ q \ \ \bigg) \bigg)
\end{gather*}

\subsection{Cycle Elimination}

\begin{figure}[t]
        \centering
            \includegraphics[scale=0.5]{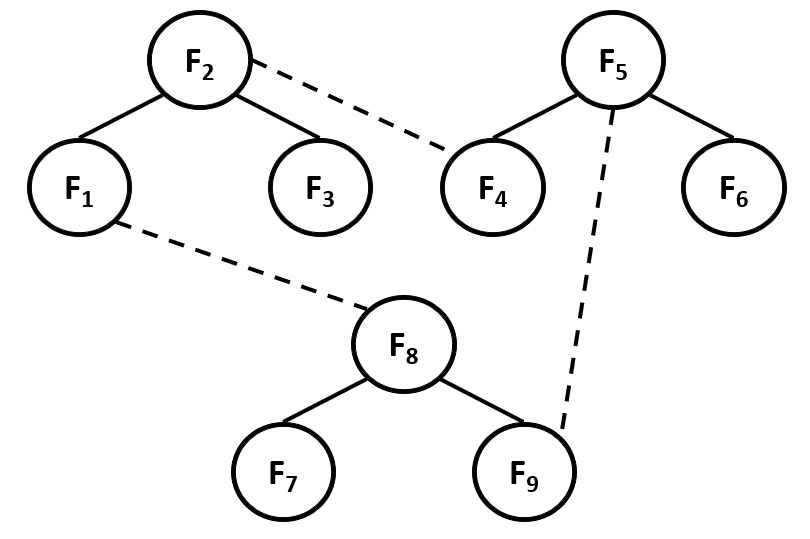}
            \caption{A cycle that leads to malformed expressions}
            \label{cycle}
\end{figure}

The aforementioned set of constraints is necessary but not sufficient to guarantee a valid graph will be produced. In particular, the constraints do not prevent cycles from occurring in reconstructed graphs (Figure \ref{cycle} provides an example). To avoid this, we enumerate all simple cycles (using the algorithm of \cite{johnson1975finding}) from the directed graph formed by parallel connectivity constraints, then assert all cycles as nogoods, i.e. the negation of a conjunction of variables. For instance, if in the set of implications a cycle was found: $(r_{i} \Rightarrow r_{j}) \wedge (r_{j} \Rightarrow r_{k}) \wedge (r_{k} \Rightarrow r_{i})$, the following negated conjunction would be asserted $\neg (r_{i} \wedge r_{j} \wedge r_{k})$. We define $\mathcal{N}$ to be the conjunction of all nogoods.

\subsection{Reconstruction}
The formula that is passed to the SAT-solver is then
\begin{gather*}
\mathlarger{\mathlarger{\mathcal{P} \wedge \mathcal{A} \wedge \mathcal{S} \wedge \mathcal{L} \wedge \mathcal{C} \wedge \mathcal{PC} \wedge \mathcal{N}}}
\end{gather*}

To see that the satisfiability problem defined above contains the input $G$ in its set of solutions, we note the following observations. First, each of $\mathcal{P}, \mathcal{A}, \mathcal{S}, \mathcal{L}$, and $\mathcal{C}$ only rule out solutions where a component of some constraint, e.g. $F_k \in \gamma_{a}$, is associated to more than one node in $G$. Clearly, because each node is associated with only one symbol, one ordered or unordered node constraint, and one parent constraint, each component to a constraint will also be associated to at most one node in $G$. Second, parallel connectivity constraints only rule out solutions where the child of some node constraint, i.e. $\gamma_{la}$, is used but the lead element of that constraint, i.e. $\gamma_{lp}$, is not used. This, by construction, cannot happen with $G$ because such tuples are not valid and are discarded (as discussed in the beginning of the section). Lastly, the inclusion of $\mathcal{N}$ only removes from the set of solutions those graphs that contain cycles consisting of only internal nodes. Because the inputs to our approach are only graphs without cycles, the inclusion of $\mathcal{N}$ cannot restrict $G$ from being a possible solution to the problem. 

\section{Additional Methods}

\subsection{Determining Structural Similarity}

At first glance, it may be tempting to make similarity judgments over the $|S| + \sum_i |C_i|$ dimensional vectors with a simple dot product or cosine similarity. However, doing so would lose much of the benefit gained by the use of multiple, parallel sets of constraints. Instead, we define an irregular matrix with one row for each of the parallel sets of constraints and one row for the set of symbols. Let $\boldsymbol{\phi}$ be a similarity measurement (e.g. dot-product, cosine similarity, etc.) that can be applied to two real-valued vectors. To compare two such matrices, $M$ and $N$, representing graphs $g$ and $h$, one first applies the comparison row-wise, i.e.

{\centering
$v = \bigg< \boldsymbol{\phi}(M_{1}, N_{1}) , \boldsymbol{\phi}(M_{2}, N_{2}), \ldots, \boldsymbol{\phi}(M_{1 + |C|}, N_{1 + |C|}) \bigg>$
\par }


The similarity of $g$ and $h$ is then given by the $\min$ of $v$, i.e.

{\centering
$\mathcal{S}_{\phi}(g, h) = \min_i v_i$
\par }


As each row of the matrix specifies an independent decomposition of a given input, the overall similarity between two matrices can be no larger than the similarity between any pair of parallel rows. This mirrors continuous logic \cite{ben2010continuous}, where the $\min$ operation is used as a conjunction, which, in our case, is used to ensure similarity across all sets of constraints.

\subsection{Masking}

The constraint encoding method as defined above has difficulty reconstructing graphs that have several identical internal nodes. To overcome this difficulty, one may either define deeper constraints, or masks can be applied to internal nodes that reduce the likelihood of encountering identicality in an input. For instance, one can define a depth mask, which extends a node's label with the node's depth in the source graph. Additionally, one might specify an argument number mask, which extends the label of a non-leaf node (there is no benefit to masking leaf nodes) with its place in the argument list of its parent node. To incorporate these masks, one simply extends $\Omega$ (which then results in an extension of $S$) with the possible masks a node could be given.

\begin{figure}[t]
\centering
            \includegraphics[scale=0.5]{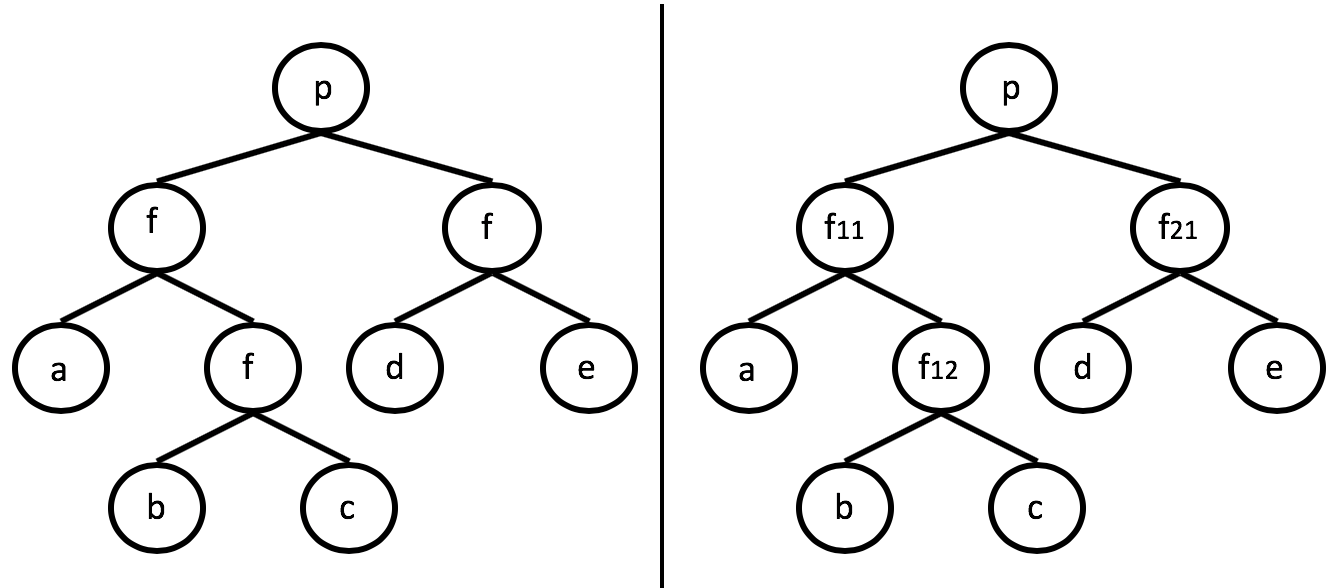}
            \caption{On the left is a tree that is ordinarily difficult to represent, while on the right is the equivalent extended tree with argument number and depth masks applied to internal nodes}
            \label{diff}
\end{figure}

\section{Experiments and Results}

Experiments were performed in two domains we believed would benefit from this approach, natural language processing and automated reasoning. In all experiments, we defined a placeholder signature \textit{a priori}, from which the constraints were generated. During testing, symbols were assigned to placeholders when they were first encountered.

In the tables of results, $|S|$ will refer to the size of the placeholder signature, $\sum_i |C_i|$ will refer to the size of the constraint sets and $t$ will refer to the number of parallel sets of constraints. In the automated reasoning experiments, our method was set to apply argument number masks to every input. CryptoMiniSAT \cite{DBLP:conf/sat/SoosNC09} was used as the underlying SAT solver for all experiments.

\subsection{Reconstructing Natural Language Sentences}

\begin{table}[b]
\caption{Results for sentence reconstruction}
\label{tsr}
\begin{center}
\begin{small}
\begin{sc}
\begin{tabular}{c|ccc|c}
\toprule 
 $t$ & $|S|$ & $\sum_i |C_i|$ & $|S| + \sum_i |C_i|$ & Corr. \\ 
 \midrule
1 & 20150 & 1495 & 21645 & 15.5\% \\ 
2 & 20150 & 2990 & 23140 & 71.5\% \\ 
3 & 20150 & 4485 & 24635 & 98.9\% \\ 
4 & 20150 & 5980 & 26130 & 99.9\% \\ 
5 & 20150 & 7475 & 27625 & 100.0\% \\
\bottomrule
\end{tabular}
\end{sc}
\end{small}
\end{center}
\end{table}

To test sentence reconstruction accuracy, we extracted 5,000 sentences (selected at random) from the Brown Corpus and translated them into ordered binary trees, where each leaf node is either a word or an end of sequence token. The average length of the extracted sentences was 22 tokens, with a median sentence length of 19 tokens, and a maximum sentence length of 147 tokens. The signature consisted of a set of 20,000 constants to cover the full set of tokens, a single predicate $\{f_1\}$, and 149 functions (i.e. $\{f_2, \ldots, f_{150}\}$, one for each word. The $split$ function was called with a width $w = 5$ for all experiments. A single test consisted of taking a sentence, converting it into a vector, and then converting that vector back to a sentence. If the reconstructed sentence matched the original sentence perfectly, it was counted as a success. All tests were given 5 seconds.

Results are shown in Table \ref{tsr}. As can be seen from the table, reconstruction accuracy increases very quickly with $t$. As a loose justification, consider that for any two words, the likelihood that both words are confused with each other is roughly the likelihood that both words are in $2t$ constraints ($t$ constraints for both of their positions in the sentence). A back-of-the-envelope calculation of the collision probability of any two words in a sentence gives $w^{-2t}$.

\subsection{Reconstructing First-Order Logic Clauses}

\begin{figure}[t]
\centering
    \begin{minipage}{0.45\textwidth}
        \centering
            \includegraphics[scale=0.5]{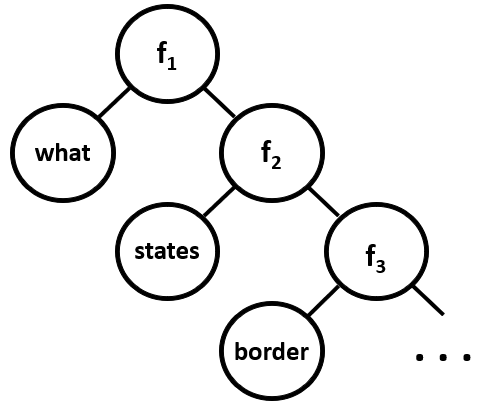}
            \caption{The binary tree form of "What states border Texas?"}
            \label{grbsent}
    \end{minipage}\hfill
    \begin{minipage}{0.45\textwidth}
        \centering
            \includegraphics[scale=0.5]{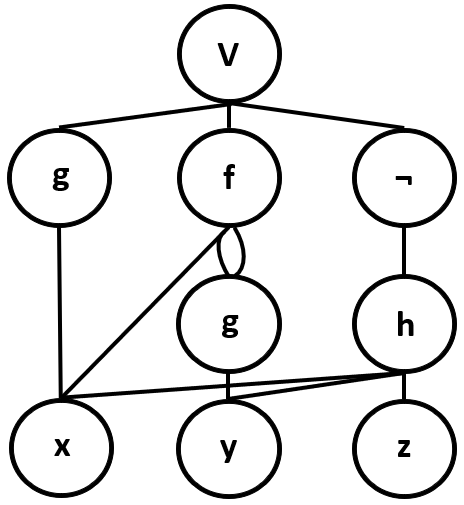}
            \caption{$g(x) \vee f(x , g(y), g(y)) \vee \neg h(x, y, z)$ in graph form}
            \label{grrep}
    \end{minipage}
\end{figure}

To test clause reconstruction, we extracted 5229 axioms in conjunctive normal form from the TPTP dataset \cite{Sut17}. Figure \ref{grrep} shows the graph of a clause. The maximum number of parents for all nodes and arity for unordered nodes was set to 5, and the maximum arity for ordered nodes was 3. These caps were chosen to reduce the memory footprint. The width for parent and unordered node constraints was 4 and for ordered node constraints was 5.

To handle variables, we first note that a clause in conjunctive normal form contains only universally quantified variables. We introduced $n$ new constants (e.g. $\{ var_1, \ldots, var_n \}$) to $\Sigma$ (variable placeholders). Then, any clause could be replaced with a logically equivalent alphabetic variant using some subset of the $n$ variable placeholders. Those nodes with a label in $\{ =, \neq, \vee \}$ were considered unordered nodes, and all other nodes were considered ordered.

\begin{table}[b]
\caption{Results for clause reconstruction}
\label{csr}
\begin{center}
\begin{small}
\begin{sc}
\begin{tabular}{c|cc|ccc}
\toprule 
$t$ & $|S|$ & $\sum_i |C_i|$ & Corr. & Incorr. & Timeout \\
\midrule
1 & 2000 & 4818 & 48.6\% & 25.6\% & 9.8\% \\ 
2 & 2000 & 9636 & 71.8\% & 5.3\% & 6.8\%\\ 
3 & 2000 & 14454 & 70.8\% & 2.1\% & 11.1\%\\ 
\bottomrule
\end{tabular}
\end{sc}
\end{small}
\end{center}
\end{table}

Lastly, given the ubiquity and importance of negations applied to non-equality predicates, we had the constraint generation procedure separate the negation symbol into its own set, e.g. operating over $split(w, \Omega \setminus \{ \neg \}) \cup \{ \{ \neg \} \}$ rather than $split(w, \Omega)$. Argument number masks bypassed negation nodes, instead applying to the argument of a negation node (negation nodes have only one argument). A reconstructed clause was considered correct if its string representation exactly matched the string of the original clause. All clause reconstructions were given 30 seconds to complete.

Results are shown in Table \ref{csr}. Of the 5229 clauses our method was given, 839 (i.e. 16.0\%) had at least one node with too many children or too many parents, and were thus unable to be represented (these were simply counted as failures). As can be seen from the table, the failure modes for $t = 2$ and $t = 3$ were somewhat contrasting. We suspect that an implementation of the satisfiability problem generator in a lower level language may solve the issue of timeouts, as actually solving the satisfiability problem never contributed substantially to the runtime of our approach.

\subsection{Clause Categorization}

\begin{figure}[t] 
\begin{center}
\includegraphics[scale=0.4]{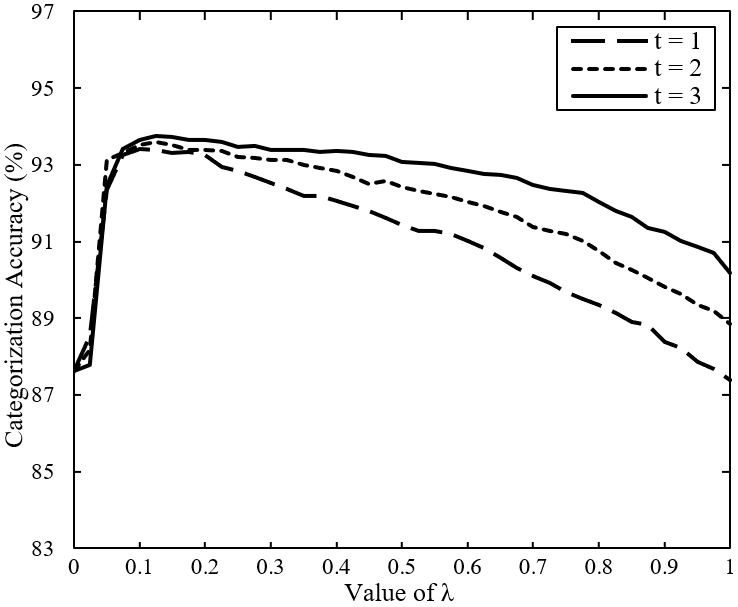}
\end{center}
\caption{Similarity measurement comparisons}
\label{knn}
\end{figure}

To see whether these representations could provide any utility beyond bag-of-words representations, we explored their use for clause categorization. Starting from the same 5229 clauses used for the reconstruction experiments, we first removed all clauses not representable by the same graph decomposition code used for the previous experiment (i.e. the 839 clauses containing a node with too many parents or children). We then processed each clause to replace all variables with a single consistent token and removed all duplicates from that set, leaving 4210 clauses to use for experiments. The conversion of a clause into a vector representation was performed with the same decomposition code used for the reconstruction experiments.

Given a clause, the task was to determine its source theory in the TPTP (e.g. the label RNG for a clause in the Ring Theory domain). We wrote a simple 1-nearest neighbor classifier that would categorize a test clause with the same category as the most similar example from the training data (we found performance with both bag-of-words and our representations degraded for larger neighborhoods, thus, we only report results for the 1-NN classifier). We explored similarity as a combination of structural and symbol similarities. That is, letting $\mathcal{B}_{cos}(g, h)$ be the cosine similarity between bag-of-words representations of graphs $g$ and $h$, we defined the similarity of $g$ and $h$ to be $\lambda \mathcal{S}_{cos}(g, h) + (1 - \lambda)\mathcal{B}_{cos}(g, h)$. Different values of $\lambda \in [0, 1]$ then led to different emphases on either structural or symbol similarity (with $\lambda = 0$ being purely symbol, i.e. bag-of-words, similarity and $\lambda = 1$ being purely structural similarity).

The average results for five-fold cross validation can be seen in Figure \ref{knn}. As might be expected, more parallel sets of constraints, i.e. higher $t$ values, led to better performance. Purely symbol similarity led to 87.6\% performance, purely structural similarity (at $t = 3$) led to 90.2\% performance, but a combination of the two (that leaned heavily towards symbol similarity) led to the highest performance at 93.8\%.

\section{Conclusions and Future Work}

In this paper, we introduced a method for producing high fidelity, decodable, fixed-length vectors of structured data. We see two promising avenues for immediate future work. The first is to extend this method to handle undirected graphs. Second, the vectors presented in this work capture only coarse-grained, discrete similarity. For fine-grained similarity, these representations will likely to be more compositional and support continuous similarity.

\bibliography{template.bbl}
\bibliographystyle{unsrt}

\end{document}